\documentclass[runningheads]{llncs}
\usepackage{graphicx}
\usepackage{hyperref}
\usepackage{comment}
\usepackage{booktabs}
\usepackage{array}
\renewcommand{\c}{\centering}
\hypersetup{
colorlinks=true,breaklinks=true,
linkcolor=blue,citecolor=blue,
urlcolor=black}

\begin{document}
\title{Label-Efficient Multi-Task Segmentation 
\\ using Contrastive Learning}
\author{Junichiro Iwasawa\inst{1}
\thanks{This work was done when J.I. worked at Preferred Networks as an intern and part-time researcher}\and
Yuichiro Hirano\inst{2} \and
Yohei Sugawara\inst{2}}
\authorrunning{J. Iwasawa {\it et al}}
\institute{The University of Tokyo, Tokyo, Japan\\\email{jiwasawa@ubi.s.u-tokyo.ac.jp}\\ \and
Preferred Networks, Tokyo, Japan\\
\email{\{hirano,suga\}@preferred.jp}}
\maketitle

\begin{abstract} 
Obtaining annotations for 3D medical images is expensive and time-consuming, despite its importance for automating segmentation tasks. 
Although multi-task learning is considered an effective method for training segmentation models using small amounts of annotated data, a systematic understanding of various subtasks is still lacking. 
In this study, we propose a multi-task segmentation model with a contrastive learning based subtask and compare its performance with other multi-task models, varying the number of labeled data for training.
We further extend our model so that it can utilize unlabeled data through the regularization branch in a semi-supervised manner.
We experimentally show that our proposed method outperforms other multi-task methods including the state-of-the-art fully supervised model when the amount of annotated data is limited. 

\end{abstract}

\section{Introduction}
For precision medicine, it is imperative that interpretation and classification of medical images is done quickly and efficiently; however, it is becoming a major hurdle due to the shortage of clinical specialists who can provide informed clinical diagnoses.
Automated segmentation can not only save physicians' time but can also provide accurate and reproducible results for medical analysis.
Recent advances in convolutional neural networks (CNN) have yielded state-of-the-art segmentation results for both 2D and 3D medical images \cite{unet2015,vnet2016,myronenko18}, a significant step toward fully automated segmentation. 
However, this level of performance is only possible when sufficient amount of labeled data is available.
Furthermore, obtaining annotations from medical experts is both expensive and time-consuming, despite its importance for training CNNs.
Thus, methods that utilize small labeled datasets have been explored extensively. 
Specifically, multi-task learning has been considered as an efficient method for small data, since parameter sharing for both the main segmentation task and regularization subtask could reduce the risk of overfitting \cite{ruder2017,myronenko18}.
Although subtasks in multi-task learning have been extensively investigated, we still lack a systematic understanding of the impact of subtasks on the main segmentation model, especially in the low labeled data regime. 

Contrastive learning based approaches have recently shown state-of-the-art performance in image classification with small amount of labels \cite{oord2018,cpc2019,moco2019,chen2020}.
In this work, we integrated a contrastive learning based approach as a subtask of a multi-task segmentation model.
However, its applicability in segmentation tasks, especially medical image segmentation is yet to be explored \cite{moco2019}.
Here, we systematically assess the performance of a brain tumor segmentation problem for three different multi-task models including our proposed method.
The main contributions can be summarized as follows\footnote{Our implementation is available at \href{https://github.com/pfnet-research/label-efficient-brain-tumor-segmentation}{\color{blue}github.com/pfnet-research/label-efficient-brain-tumor-segmentation}.}:
\begin{itemize}
    \item We propose a novel method for tumor segmentation by utilizing contrastive learning as a subtask for the main segmentation model.
    \item We experimentally show that our proposed method, combined with a semi-supervised approach which utilizes unlabeled data, could enhance segmentation performance when the amount of labeled data is small.
\end{itemize}

\begin{figure}[t]
\centering
\includegraphics[width=\textwidth]{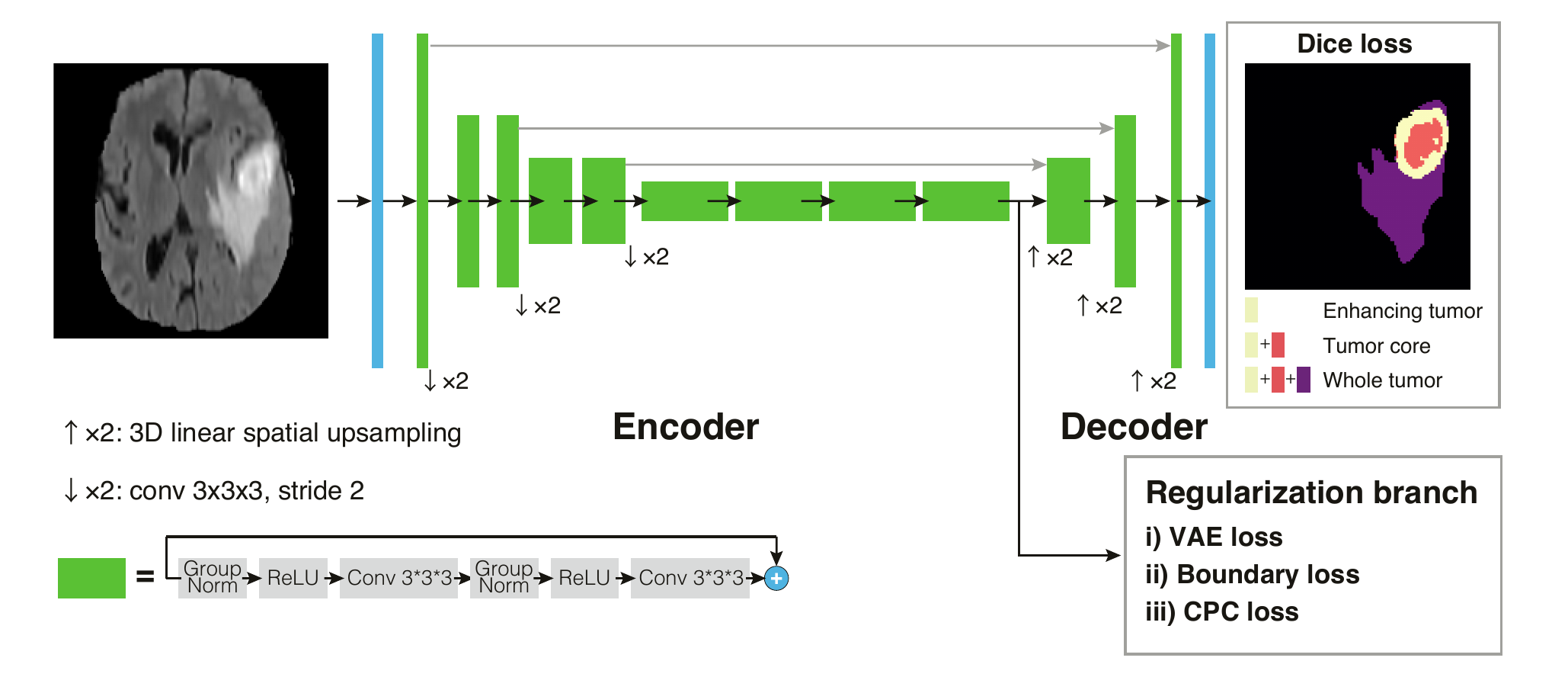}
\caption{Schematic image of the encoder-decoder based network combined with a regularization branch. The green block represents a ResBlock.} \label{fig1}
\end{figure}

\section{Methods}

\subsection{Encoder-decoder network with regularization branches}

Fig. \ref{fig1} is the schematic image of our model. 
The model constitutes an encoder-decoder architecture with a regularization branch starting from the encoder output. 
The encoder and decoder are composed of ResNet \cite{resnet2016}-like blocks (ResBlock) with skip-connections between the encoder and decoder \cite{unet2015,vnet2016,myronenko18}. 
We adopted the encoder-decoder architecture from Myronenko (2018) \cite{myronenko18}, and each ResBlock consists of two $3\times3\times3$ convolution layers with group normalization \cite{groupnorm}. 

The loss function of the model consists of two terms: $\mathcal{L}_{\rm total} = \mathcal{L}_{\rm Dice} + \mathcal{L}_{\rm branch}$,
where $\mathcal{L}_{\rm Dice}$ is a softmax Dice loss applied to the decoder output $p_{\rm pred}$, compelling it to match the ground truth label $p_{\rm true}$. The Dice loss \cite{vnet2016} is given by,
\begin{equation}
    \mathcal{L}_{\rm Dice}\left(p_{\rm true},p_{\rm pred}\right) =
    1 - \frac{2\times\sum_i p_{{\rm true},i}p_{{\rm pred},i}}{\sum_i (p_{{\rm true},i}^2 +p_{{\rm pred},i}^2) + \epsilon},
\end{equation}
where $\epsilon=10^{-7}$ is a constant preventing zero division. 
$\mathcal{L}_{\rm branch}$ is the loss applied to the output of the regularization branch and is dependent on the architecture of the branch. 

To explore the regularization effects of the multi-task learning, we compared the performance of different subtasks for the regularization branch on the encoder endpoint. 
Based on the type of information it uses to calculate the loss function, subtasks for multi-task learning can be categorized into the following classes:

\begin{enumerate}
    \item Subtasks that use the input $X$ itself, without using any additional information, such as a decoder-like branch attempting to reconstruct the original input $X$ \cite{myronenko18}. 
    \item Subtasks that attempt to predict a transformed feature of the label $y$, such as boundary-aware networks that predict the boundary of the given label \cite{myronenko2019,myronenko2019b}.
    \item Subtasks that compel the encoder to obtain certain representations of the input by predicting low or high-level features from the input $X$, such as tasks predicting the angles of rotated images \cite{noroozi2016}. 
    Our proposed method using contrastive predictive coding (CPC) \cite{oord2018,cpc2019} would also be classified in this class.
\end{enumerate}

To investigate a wide spectrum of subtasks, we implemented three different types of regularization branches that use either a variational autoencoder (VAE) loss, a boundary loss, or a CPC loss. The description of each branch and loss is provided subsequently (the architectures are shown in supplementary materials).

\subsubsection{Variational autoencoder branch}
The purpose of the VAE branch is to guide the encoder to extract an efficient low-dimensional representation for the input \cite{myronenko18}.
The encoder output is first fed to a linear layer to reduce its dimension to 256 (128 to represent the mean $\mu$, and 128 to represent the standard deviation (SD) $\sigma$). 
Accordingly, a sample is drawn from a 128-dimension Gaussian distribution with the given mean and SD. 
This sample is processed by several upsizing layers, where the number of features is reduced by a factor of two using $1\times1\times1$ convolution, and the spatial dimension is doubled using 3D bilinear sampling, so that the final output size matches the input size. 
The output is fed to the VAE loss ($\mathcal{L}_{\rm VAE}$) which is given by $\mathcal{L}_{\rm branch} = \mathcal{L}_{\rm VAE} = 0.1\times \left(\mathcal{L}_{\rm rec}+\mathcal{L}_{\rm KL} \right)$, 
where $\mathcal{L}_{\rm rec}$ is the L2 loss between the output of the VAE branch and the input image, and $\mathcal{L}_{\rm KL}$ is the Kullback-Leibler divergence between the two normal distributions, $\mathcal{N}(\mu,\sigma^2)$ and $\mathcal{N}(0,1)$.

\subsubsection{Boundary attention branch}
The boundary attention branch aims to regularize the encoder by extracting information for predicting the boundaries of the given labels \cite{attention2017,myronenko2019,myronenko2019b}. 
We prepared the boundary labels by applying a 3D Laplacian kernel to the ground truth binary labels. The attention layer first upsizes the output of the encoder, and concatenates it with the feature map from the encoder with the same spatial dimensions. 
Accordingly, a $1\times1\times1$ convolution is applied to the concatenated sample, followed by a sigmoid function that yields an attention map. 
The output of each attention layer is given by an element-wise multiplication of the attention map and the input to the layer. 
This operation is repeated until the spatial dimension matches that of the model input.

The loss for the boundary attention branch is given by $\mathcal{L}_{\rm branch} = \mathcal{L}_{\rm boundary} = \mathcal{L}_{\rm Dice}\left(b_{\rm pred},b_{\rm true} \right) + \mathcal{L}_{\rm edge}$,
where $\mathcal{L}_{\rm Dice}\left(b_{\rm pred},b_{\rm true} \right)$ is the Dice loss between the branch output ($b_{\rm pred}$) and boundaries of the ground truth label ($b_{\rm true}$). $\mathcal{L}_{\rm edge}$ is given by a weighted binary cross entropy loss:
\begin{equation}
    \mathcal{L}_{\rm edge} = 
    -\beta\sum_{j\in y_+}\log{P\left(b_{{\rm pred},j}=1|X,\theta\right)} 
    -(1-\beta)\sum_{j\in y_-}\log{P\left(b_{{\rm pred},j}=0|X,\theta\right)},
\end{equation}
where $X$, $\theta$, $y_+$ and $y_-$ denote the input, model parameters, and the boundary and non-boundary voxels, respectively. $\beta$, the ratio of the non-boundary voxels to the total number of voxels is introduced to handle the imbalance of the boundary and non-boundary voxels.

\subsubsection{Contrastive predictive coding branch}
Self-supervised learning, where a representation is learned from unlabeled data by predicting missing input data based on other parts of the input, has become a promising method for learning representations; it is useful for downstream tasks such as classification \cite{kolesnikov2019,cpc2019} and segmentation \cite{modelsgenesis}. 
Recently, CPC has been proposed as a self-supervised method that can outperform the fully supervised methods in ImageNet classification tasks in the small labeled data regime \cite{cpc2019}. 
Despite its performance in classification tasks, the effectiveness of CPC in segmentation tasks is yet to be explored. 
Here, we incorporated the CPC architecture into the encoder-decoder structure as a regularization branch and investigated its performance with regard to medical image segmentation.

First, we divided the input image into $32\times32\times32$ overlapping patches with a 16-voxel overlap (resulting in an $8\times8\times7$ grid for the brain tumor dataset). 
Each divided patch is individually encoded to a latent representation using the encoder in the encoder-decoder architecture and spatially mean-pooled into a single feature vector $z_{i,j,k}$. 
Here, it should be noted that the visual field of the encoder when using the CPC branch is $32\times32\times32$ pixels, which is smaller than that with the other regularization branches due to the initial image division.
Accordingly, eight layered ResBlocks $f_{\rm res8}$ and a linear layer $W$ were applied to the upper half of the feature vectors, $\hat{z}_{i,j,k_{\rm low}} = W\left(f_{\rm res8}\left(z_{i,j,k_{\rm up}}\right)\right)$, to predict the lower half of the feature vectors $z_{i,j,k_{\rm low}}$. 
The predictions were evaluated based on the CPC loss (i.e. the InfoNCE \cite{oord2018}),
\begin{equation}
    \mathcal{L}_{\rm branch} = \mathcal{L}_{\rm CPC} = -\sum_{i,j,k_{\rm low}}\log{
    \frac{\exp{\left(\hat{z}^T_{i,j,k_{\rm low}}z_{i,j,k_{\rm low}}\right)}}
    {\exp{\left(\hat{z}^T_{i,j,k_{\rm low}}z_{i,j,k_{\rm low}}\right)}+\sum_l \exp{\left(\hat{z}^T_{i,j,k_{\rm low}}z_{l}\right)}}
    },
\end{equation}
where the negative samples $\{z_{l}\}$ are randomly taken from the other feature vectors which are encoded from different patches of the same image.

\begin{figure}[t]
\centering
\includegraphics[width=\textwidth]{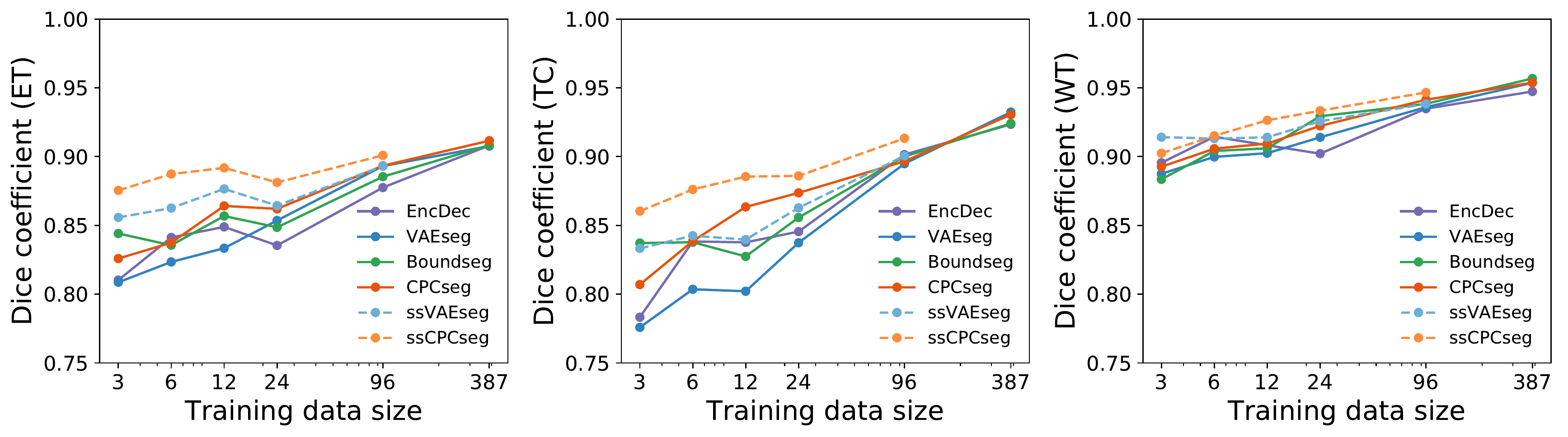}
\caption{Comparison of the Dice scores of the test predictions by the proposed models with different training data sizes. For the ssVAEseg and ssCPCseg, the training data size refers to the number of labeled images.} \label{fig2}
\end{figure}

\section{Experiments and Results}

\subsubsection{Brain tumor dataset}
We used the brain tumor dataset provided by the Medical Segmentation Decathlon \cite{MSD2019}, which is a subset of the data used in the 2016 and 2017 Brain Tumor Image Segmentation (BraTS) challenges \cite{brats2015,brats2017,brats2018}.
This dataset includes multimodal 3D magnetic resonance imaging (MRI) scans from patients diagnosed with either glioblastoma or lower-grade glioma with an image size of $240\times240\times155$. 
For the experiments, we randomly split the 484 labeled data into the training (387), validation (48), and test set (49). 

The dataset has three different labels corresponding to different tumor sub-regions (i.e., necrotic and non-enhancing parts of the tumor (NCR \& NET), the peritumoral edema (ED), and the enhancing tumor (ET)). 
Following the BraTS challenge, we evaluated the model's prediction accuracy using the three nested structures of these sub-regions: ET, tumor core (TC: ET+NCR+NET), and the whole tumor (WT: TC+ED). 
The output of the network was set to three channels to predict each of the three nested structures above.

\subsubsection{Preprocessing and augmentation}
All the input images were normalized to have zero mean and unit SD. Accordingly, a random scale (0.9, 1.1), random intensity shift (-0.1, 0.1 of SD), and random axis mirror flip (all axes, probability 0.5) were applied.
All the inputs were randomly cropped to $(160, 192, 128)$, except when utilizing the CPC branch ($(144,144,128)$ in this case), due to the graphics processing unit (GPU) memory limitation. Note that the random cropping was performed without respect to the location of the brain region.

\begin{table}[b]
\centering
\caption{Performance of EncDec, VAEseg, Boundseg, CPCseg (trained with six labeled data), and ssVAEseg, ssCPCseg (trained with six labeled and 381 unlabeled data). Evaluation was done using the mean Dice and $95^{\rm th}$ percentile Hausdorff distance. The performance of VAEseg trained with 387 labels is shown for comparison.}\label{dice_hausdorff}
\begin{tabular}[width=\textwidth]{|p{3cm}|>{\c}p{1.4cm}|>{\c}p{1.4cm}|>{\c}p{1.4cm}|>{\c}p{1.4cm}|>{\c}p{1.4cm}|c|}
\hline
& \multicolumn{3}{c}{Dice} & \multicolumn{3}{|c|}{Hausdorff distance (mm)}
\\
\hline
Test data & ET &TC &WT &ET & TC &WT \\
\hline
VAEseg (387 labels) & 0.9077 & 0.9323 & 0.9536 & 3.6034 & 10.2344 & 8.3895 \\
\hline
EncDec (6 labels) & 0.8412 & 0.8383 & 0.9144 & 11.4697 & 20.12 & 24.3726 \\
VAEseg (6 labels) & 0.8234 & 0.8036 & 0.8998 & 14.3467 & 22.4926  & 17.9775 \\
Boundseg (6 labels) & 0.8356 & 0.8378 & 0.9041 & 17.0323 & 27.2128 & 25.8112\\
CPCseg (6 labels) & 0.8374 & 0.8386 & 0.9057 & 10.2839 & 14.9661 & 15.0633 \\
ssVAEseg (6 labels) & 0.8626 & 0.8425 & 0.9131 & 9.1966 & {\bf 12.5302}  & 14.8056 \\
ssCPCseg (6 labels) & {\bf 0.8873} & {\bf 0.8761} &{\bf 0.9151} &{\bf 8.7092} &16.0947 &\ {\bf 12.3962}\ \ \\
\hline
\end{tabular}
\end{table}

\subsubsection{Comparison of different regularization branches}
We implemented our network in Chainer \cite{chainer2019} and trained it on eight Tesla V100 32GB GPUs. 
We used a batch size of eight, and the Adam optimizer \cite{adam} with an initial learning rate of $\alpha_0 = 10^{-4}$ that was further decreased according to $\alpha=\alpha_0(1-n/N)^{0.9}$, where $n,N$ denotes the current epoch and the total number of epochs, respectively.

To compare the different regularization effects, we measured the segmentation performances of four models: the encoder-decoder alone (EncDec), EncDec with a VAE branch (VAEseg), EncDec with a boundary attention branch (Boundseg), and EncDec with a CPC branch (CPCseg). 
The performance was evaluated using the mean Dice and $95^{\rm th}$-percentile Hausdorff distance of the ET, TC, and WT. 
To evaluate the regularization effect with varied amounts of labeled data, we also evaluated each model's performance with the training data size reduced to 3 -- 96. The results are given in Fig. \ref{fig2} and Table \ref{dice_hausdorff}, \ref{full_dice_table}.

It can first be observed that no regularization branch consistently outperforms the others. 
Furthermore, it can be observed that in some cases the EncDec that has no regularization branch had the highest mean Dice score among the fully supervised models (Table \ref{dice_hausdorff}). 
However, it should also be noted that multi-task models tended to outperform the EncDec when using all 387 labeled data (Table \ref{full_dice_table}).
These results imply that the regularization branches using labeled data have a limited effect on the segmentation performance when the amount of labeled data is small.
Interestingly, the VAEseg, the state-of-the-art model in the BraTS 2018 challenge \cite{myronenko18}, was not necessarily the best model for various training data sizes.
This was surprising, although the dataset we used in this study slightly differs from that used in the BraTS 2018 challenge.
Our results suggest that the VAE subtask would not always be the optimal approach to brain tumor segmentation tasks.

\subsubsection{Semi-supervised multi-task learning}
\begin{figure}[t]
\centering
\includegraphics[width=\textwidth]{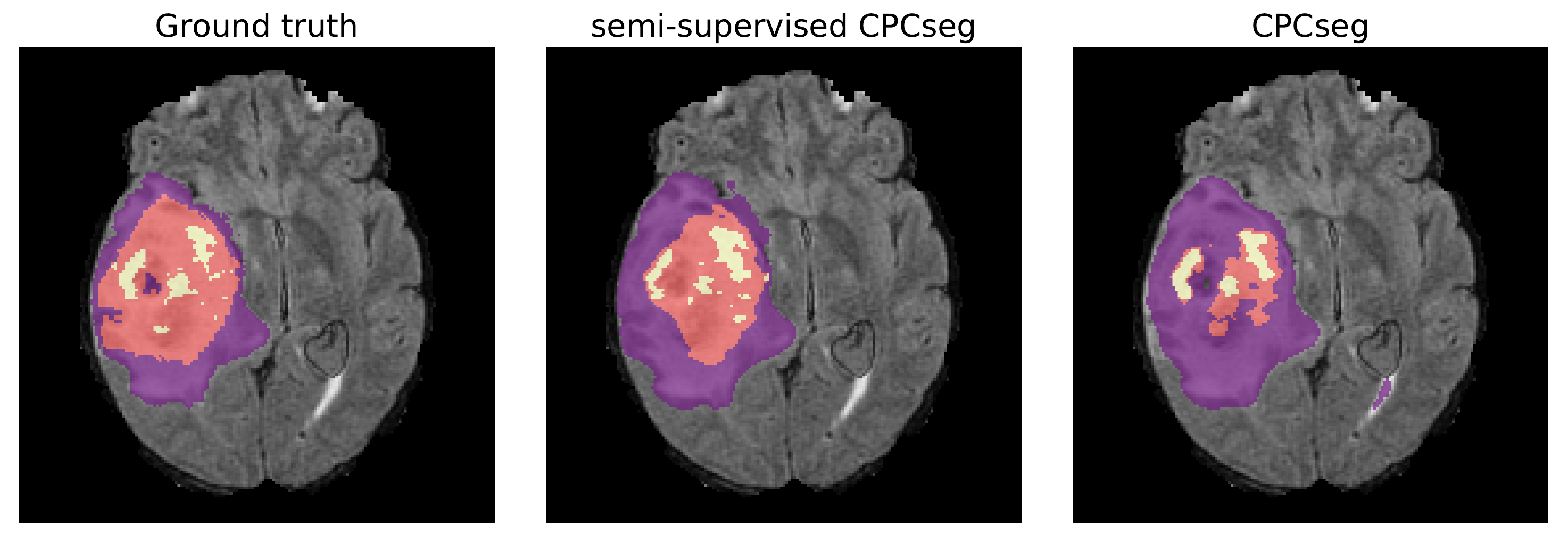}
\caption{Prediction results of the semi-supervised CPCseg (24 labeled + 363 unlabeled) and the CPCseg (24 labeled). The union of purple, orange, and yellow corresponds to the WT, orange plus yellow corresponds to the TC, and yellow corresponds to the ET.} \label{fig3}
\end{figure}

Typically, unlabeled data is readily accessible in large quantities, compared to annotated data. 
This naturally leads to the question: is it possible to utilize unlabeled data to guide the segmentation model at small data regimes? 
To answer this question, we focused on the VAEseg and CPCseg, because they do not require labels to optimize the regularization branch. 
For training, we used all the 387 unlabeled data from the training set, and varied the number of labels used. 
To utilize the unlabeled data, we devised a semi-supervised update method wherein the model could be updated using only $\mathcal{L}_{\rm branch}$ when the image had no label, and by $\mathcal{L}_{\rm Dice}+\mathcal{L}_{\rm branch}$, otherwise. 
This update method lets the encoder (and regularization branch) learn representations from both labeled and unlabeled data.
The segmentation results for the semi-supervised VAEseg (ssVAEseg) and semi-supervised CPCseg (ssCPCseg) are shown in Fig. \ref{fig2}, \ref{fig3} and Table \ref{dice_hausdorff}. 
It can be observed that the semi-supervised methods outperform their fully supervised counterparts. 
In addition, the ssCPCseg outperformed all the other regularization methods, including the fully supervised state-of-the-art model, VAEseg, in the small labeled data regime. 
This tendency was most apparent for the ET and TC. 
For example, the ssCPCseg using six labels exhibited a 6\% decrease in the Dice score for TC compared to the VAEseg using 387 labels, while other methods using six labels exhibited a 10 -- 14\% decrease (Table \ref{dice_hausdorff}).
We speculate that this is because the areas of the ET and TC are smaller, compared to that of the WT, and thus, provide less supervision signals per sample to the model.
Our results imply that difficult and non-trivial subtasks such as CPC, as well as unlabeled data, can be exploited to achieve state-of-the-art performance when the amount of annotated data is limited.

\begin{figure}[t]
\centering
\includegraphics[width=0.9\textwidth]{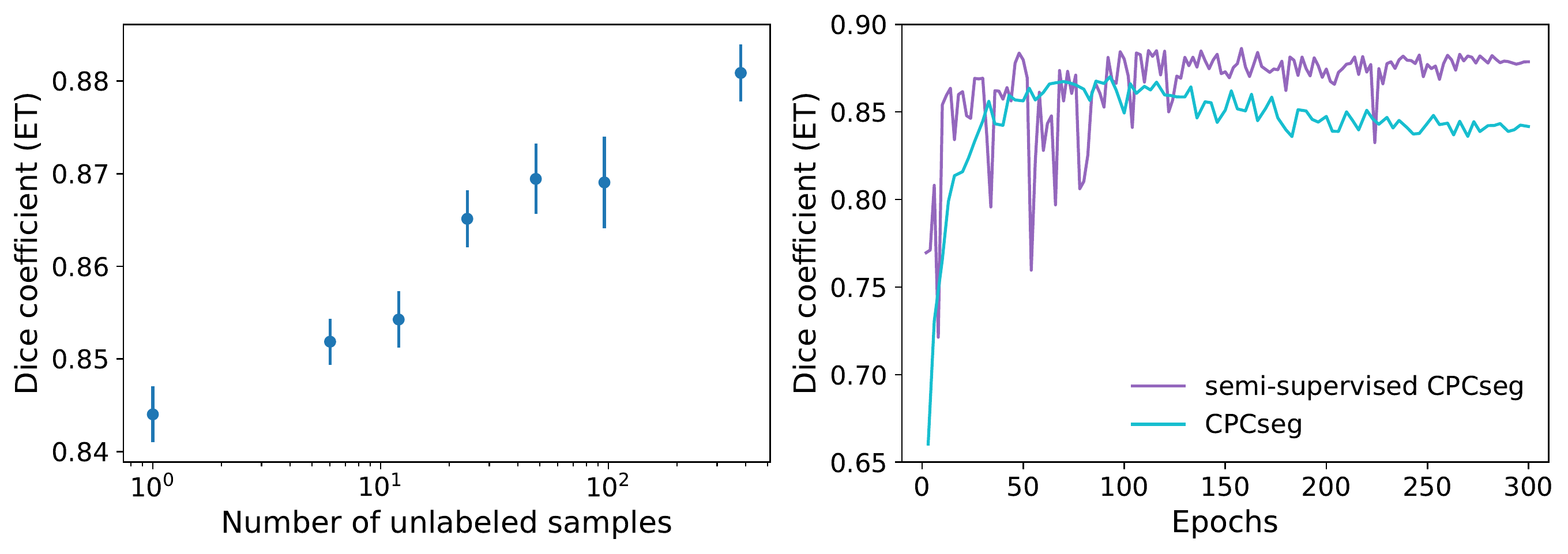}
\caption{The left panel shows the dependency of the number of unlabeled data on the segmentation accuracy for the ssCPCseg when using six labeled data. The right panel show learning curves for the test data of the brain tumor dataset. The learning curves for the CPCseg (24 labeled) and ssCPCseg (24 labeled + 363 unlabeled) are shown.} \label{fig4}
\end{figure}

To explore the importance of the number of unlabeled data to the segmentation performance, we compared the ET Dice score of the ssCPCseg trained using six labeled and different amounts of unlabeled data (Fig. \ref{fig4}). 
It can be observed that the Dice score increases monotonically with the number of unlabeled data. 
However, it should be noted that the Dice score seems to increase linearly with the $\log$ of unlabeled data, indicating that the number of data required to improve accuracy would increase exponentially. 
We also investigated the effect of unlabeled data on the learning behavior of the model. 
As can be seen in Fig. \ref{fig4}, the semi-supervised CPCseg not only produced more accurate results but was also more robust to overfitting. 
This might be because the encoder needs to map good representations for both the labeled and unlabeled data to optimize the CPC objective. 
Overall, our results imply that utilizing unlabeled data could both enhance and stabilize the model's performance.

\section{Discussion and Conclusion}
In this work, we systematically investigated the effectiveness of different regularization subtasks for medical image segmentation.
Our experiments on the brain tumor dataset showed that utilizing unlabeled data through the regularization branch improved and stabilized the performance of segmentation models when the number of labeled data was small.
Especially, our proposed ssCPCseg outperformed other methods including the state-of-the-art fully supervised model in the small labeled data regime.
In previous works, CPC has been used for self-supervised pre-training for image classification tasks with unlabeled images \cite{oord2018,cpc2019}.
Our work is the first to show the effectiveness of CPC as a regularization subtask for image segmentation by utilizing both unlabeled and labeled images, providing a novel direction for label efficient segmentation.
It should also be noted that ssCPCseg achieved particularly higher Dice scores than the other methods for tumor sub-regions with small area size (i.e. ET and TC) whose structure varies across the data and provides limited supervision signals to the model. 
Although our results are based on a single dataset, we believe that our method could be applicable to various targets in the field since target labels for medical images are often relatively small, and have widely varied structures. 

It is generally expensive to obtain annotations for 3D medical images. 
On the other hand, large number of unlabeled images are often available. Thus, our semi-supervised method should have wide applicability to medical image segmentation tasks. 
However, it should be noted that all the MRI scans in the brain tumor dataset are normalized using a reference atlas; furthermore, they have the same size and voxel resolution. 
Therefore, we have not been able to evaluate the segmentation performance when the quality of the unlabeled data varies. 
An important future work is to evaluate the model's segmentation performance when it is fed unlabeled images from different modalities and domains. 
We believe that our systematic study provides important designing principles for segmentation models, leading to more cost-efficient medical image segmentation.

\subsubsection{Acknowledgments} 
J.I. was supported by the Grant-in-Aid for JSPS Fellows JP18J21942.

\bibliographystyle{splncs04}
\bibliography{ref}

\newpage

\section*{Supplementary Materials}
\setcounter{table}{0}
\renewcommand{\thetable}{S\arabic{table}}
\setcounter{figure}{0}
\renewcommand{\thefigure}{S\arabic{figure}}

\begin{table}
\centering
\caption{Architecture for the VAE regularizing branch. The architecture was adopted from \cite{myronenko18}. GN stands for group normalization, Conv: $3\times3\times3$ convolution, Conv1: $1\times1\times1$ convolution, AddId: addition of identity / skip connection, UpLinear: 3D linear spatial upsampling, Dense: fully connected layer. }\label{vaebranch}
\begin{tabular}[width=0.95\textwidth]{|l|c|c|}
\hline
\ Operations &  Repeats & Output size\\
\hline
\ GN, ReLU, Conv(16) stride 2, Dense (256) & 1 & $256\times1$\\
\ sample $\sim$ \begin{math}\mathcal{N}\left(\mu(128), \sigma^2(128) \right)\end{math} & 1 & $128\times1$\\
\ Dense, ReLU, Conv1, UpLinear & 1 & $256\times20\times24\times16$\\
\ Conv1, UpLinear & 1 & $128\times40\times48\times32$\\
\ GN, ReLU, Conv, GN, ReLU, Conv, AddId\  & 1 & $128\times40\times48\times32$\\
\ Conv1, UpLinear & 1 & $64\times80\times96\times64$\\
\ GN, ReLU, Conv, GN, ReLU, Conv, AddId \ & 1 & $64\times80\times96\times64$\\
\ Conv1, UpLinear \ & 1 & \ $32\times160\times192\times128$ \ \\
\ GN, ReLU, Conv, GN, ReLU, Conv, AddId \ & 1 &\ $32\times160\times192\times128$ \ \\
\ Conv1\  & 1 & $4\times160\times192\times128$\\
\hline
\end{tabular}
\end{table}

\begin{table}
\centering
\caption{Architecture for the boundary attention branch. ConcSC stands for concatenate input with skip connection, EM: element-wise multiplication with input and attention layer.}\label{boundarybranch}
\begin{tabular}[width=0.95\textwidth]{|l|c|c|}
\hline
\ Operations &  Repeats & Output size\\
\hline
\ GN, ReLU, Conv, GN, ReLU, Conv, AddId \ & 1 &\ $128\times40\times48\times32$ \ \\
\ ConcSC, Conv1, Sigmoid, EM\ & 1 & $128\times40\times48\times32$\\
\ GN, ReLU, Conv, GN, ReLU, Conv, AddId \ & 1 &\ $64\times80\times96\times64$ \ \\
\ ConcSC, Conv1, Sigmoid, EM\ & 1 & $64\times80\times96\times64$\\
\ GN, ReLU, Conv, GN, ReLU, Conv, AddId \ & 1 &\ $32\times160\times192\times128$ \ \\
\ ConcSC, Conv1, Sigmoid, EM\ & 1 & $32\times160\times192\times128$\\
\ Conv1\  & 1 & $4\times160\times192\times128$\\
\hline
\end{tabular}
\end{table}

\begin{table}[t]
\centering
\caption{Architecture for the CPC regularization branch. The initial random crop size for the model with the CPC branch is $(144, 144, 128)$. The input image is first divided into overlapping $32\times32\times32$ grids before being fed to the encoder.}\label{CPCbranch}
\begin{tabular}[width=0.95\textwidth]{|l|c|c|}
\hline
\ Operations &  Repeats & Output size\\
\hline
\ Spatial average \ & 1 &\ $1\times8\times8\times7$ \ \\
\ GN, ReLU, Conv, GN, ReLU, Conv \ & 8 &\ $1\times8\times8\times7$ \ \\
\ Conv1\  & 1 & $1\times8\times8\times7$\\
\hline
\end{tabular}
\end{table}

\begin{table}[b]
\centering
\caption{Performance of EncDec, VAEseg, Boundseg, CPCseg (trained with six and 387 labeled data, respectively), and ssVAEseg, ssCPCseg (trained with six labeled and 381 unlabeled data). Evaluation was done using the mean Dice score.}
\label{full_dice_table}
\begin{tabular}[width=\textwidth]{|p{5.5cm}|>{\c}p{1.4cm}|>{\c}p{1.4cm}|c|}
\hline
\multicolumn{4}{|c|}{using all 387 labels}\\
\hline
& \multicolumn{3}{c|}{Dice} \\
\hline
Test data & ET & TC & WT \\
\hline
EncDec (387 labels)    & 0.9081 & 0.9234 & 0.9473 \\
VAEseg (387 labels)    & 0.9077 & \textbf{0.9323} & 0.9536 \\
Boundseg (387 labels)  & 0.9083 & 0.9241 & \textbf{0.9568} \\
CPCseg (387 labels)    & \textbf{0.9116} & 0.9305 & 0.9538 \\
\hline
\multicolumn{4}{|c|}{using 6 labels}\\
\hline
EncDec (6 labels)    & 0.8412 & 0.8383 & 0.9144 \\
VAEseg (6 labels)    & 0.8234 & 0.8036 & 0.8998 \\
Boundseg (6 labels)  & 0.8356 & 0.8378 & 0.9041 \\
CPCseg (6 labels)    & 0.8374 & 0.8386 & 0.9057 \\
ssVAEseg (6 labels + 381 unlabeled)  & 0.8626 & 0.8425 & 0.9131 \\
ssCPCseg (6 labels + 381 unlabeled) &{\bf 0.8873} & {\bf 0.8761} &{\bf 0.9151} \ \\
\hline
\end{tabular}
\end{table}

\end{document}